\documentclass[letterpaper, 10 pt, conference]{ieeeconf}  

\IEEEoverridecommandlockouts                              

\usepackage{graphicx} 
\usepackage{epsfig} 
\usepackage{mathptmx} 
\usepackage{times} 
\usepackage{amsmath,amssymb,amsfonts}
\usepackage{cite}
\usepackage{pgfplots}
\pgfplotsset{compat=1.7}
\usepackage{pgfplotstable}
\usepackage{tikz,pgfplots}
\usepackage{physics} 
\usepackage{siunitx} 
\usepackage{enumerate} 
\usepackage{tikz,pgfplots}
\usepackage{verbatim}
\usepackage{color}
\usepackage{mathtools}
\usepackage{breqn}
\usepackage{subcaption}
\usepackage{array}
\usepackage{makecell}
\usepackage{soul}  
\usepackage{comment}
\usepackage{textcomp}
\usepackage{subcaption}
\usepackage{algorithm}
\usepackage{algpseudocode} 
\usepackage[labelsep=period]{caption}
\usepackage[flushleft]{threeparttable}
\usepackage[table,xcdraw,dvipsnames]{xcolor}
\usepackage{tablefootnote}
\usepackage{multirow}   
\usepackage{multicol}   
\usepackage{array}      
\usepackage{booktabs}   
\usepackage{tabularx}   
\usepackage{cite} 

\usepackage{duckuments}
\usepackage[colorlinks,citecolor=blue,urlcolor=blue,linkcolor=blue,bookmarks=false,hypertexnames=true]{hyperref} 
\usepackage{amssymb}
\usepackage{pifont}
\newcommand{\cmark}{\ding{51}}%
\newcommand{\xmark}{\ding{55}}%

\newcommand{\green}[1]{\textcolor{black}{#1}}

\newcommand{\sy}[1]{\textcolor{black}{#1}}
\newcommand{\nm}[1]{\textcolor{black}{#1}}

\title{\bf
TRIP-Bag: A Portable Teleoperation System  \\ for Plug-and-Play Robotic Arms and Leaders
}

\author{Noboru Myers$^{*}$, Sankalp Yamsani$^{*}$, Obin Kwon$^{*}$, and Joohyung Kim
\thanks{
$^{*}$These authors equally contributed to this work.
All authors are with the KIMLAB (Kinetic Intelligent Machine LAB), University of Illinois Urbana-Champaign, IL 61801, USA.  Author contact information is {\tt\small \{noborum2, yamsani2, joohyung\}@illinois.edu}}
}

\begin{document}

\maketitle


\thispagestyle{empty}
\pagestyle{empty}

\begin{abstract}
Large scale, diverse demonstration data for manipulation tasks remains a major challenge in learning-based robot policies. Existing in-the-wild data collection approaches often rely on vision-based pose estimation of hand-held grippers or gloves, which introduces an embodiment gap between the collection platform and the target robot. Teleoperation systems eliminate the embodiment gap, but are typically impractical to deploy outside the laboratory environment. We propose TRIP-Bag (Teleoperation, Recording, Intelligence in a Portable \green{Bag}), a portable, puppeteer-style teleoperation system fully contained within a commercial suitcase, as a practical solution for collecting high-fidelity manipulation data across varied settings. With a setup time of under five minutes and direct joint-to-joint teleoperation, TRIP-Bag enables rapid and reliable data collection in any environment. We validated TRIP-Bag’s usability through experiments with non-expert users, showing that the system is intuitive and easy to operate. Furthermore, we confirmed the quality of the collected data by training benchmark manipulation policies, demonstrating its value as a practical resource for robot learning.

\end{abstract}
%

\section{INTRODUCTION}
\label{section:intro}
%
\sy{Recent advances in imitation learning and generative policy models have shifted robotic manipulation toward data-driven paradigms\cite{aloha,chi2024diffusionpolicy}. However, unlike vision and language, robotics lacks access to internet-scale embodied datasets. Collecting high-fidelity, action-conditioned interaction data requires physical hardware, precise calibration, synchronized sensing, and human operation, which creates a fundamental scalability barrier. As a result, a limiting factor in robotic foundation models is the ability to collect diverse, high-quality data across environments. To overcome this data-collection bottleneck, two primary data-acquisition frameworks have gained popularity over the years}, \nm{teleoperation and handheld devices}.

\sy{Teleoperation has become a critical tool for advancing robot learning research by enabling efficient and scalable data collection.} Recent efforts have explored diverse types of teleoperation methods \cite{aloha, gello, DexPilot,hato}. However, most existing teleoperation systems are confined to laboratory environments because transporting, assembling, and calibrating platforms across sites requires extensive infrastructure and setup time. \sy{Such fixed infrastructures inherently restrict environmental diversity and interaction variability, resulting in datasets that fail to capture the distributional complexity of real-world settings.}  

To reduce the barrier towards in-the-wild large scale data, efforts have turned to handheld devices for data collection \cite{Chi_2024, Song_2020, Young_2020}. These approaches enhance portability; however, they often result in embodiment gaps between the leader and follower systems, which limit the fidelity and usability of the collected data.

\sy{This work introduces a portable teleoperation platform that combines the high-fidelity data of teleoperation frameworks with the portability of handheld device-based data collection. }
Teleoperation, Recording, Intelligence in a Portable Bag (TRIP-Bag) includes two pluggable robotic arms (PAPRAS) \cite{papras}, two pluggable scaled puppeteer leaders \cite{PAPRLE}, three RGB-D cameras, and compact computing units for teleoperation control and data collection, all contained within a commercial suitcase. 
TRIP-Bag enables researchers to collect multimodal teleoperation data in diverse settings, bridging the gap between controlled laboratory conditions and dynamic real-world environments.
\begin{figure}[t]
        \centering \includegraphics[width=\columnwidth
        ]{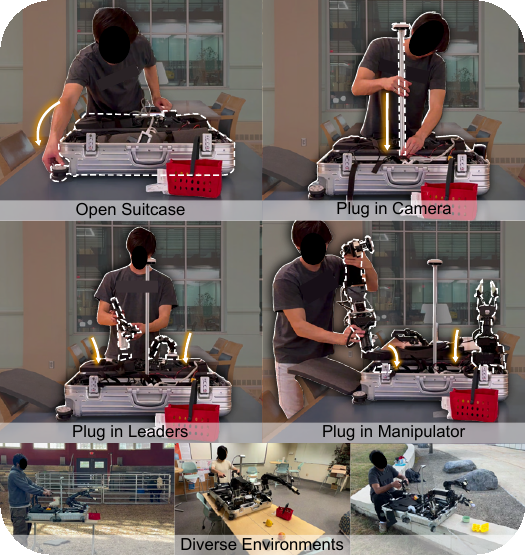}
        \caption{The setup process of collecting data with our system. A few of the diverse locations we collected in.}  
        \label{fig:main}
\end{figure}
\begin{table*}[t]
\vspace*{5pt}
\caption{\sy{Comparison of Data Collection Devices}}
\label{table:Compare}
\centering
\scriptsize
\begin{tabularx}{\textwidth}{l c c c c c c c}
\toprule
& \textbf{In-the-Wild} & \textbf{Control} & \textbf{Action-Space} & \textbf{Direct Embodiment} &  
\textbf{Calibration} & \textbf{Operator Gripper
} & \textbf{Full Proprioceptive}\\
&  & \textbf{Interface} & \textbf{Representation} & \textbf{ Mapping?} &  
\textbf{Free?} & \textbf{Feedback Mechanism} & \textbf{State Logging} \\
\midrule
UMI\cite{Chi_2024} & \cmark & Handheld & Task space & \xmark   & \xmark & Passive trigger resistance & \xmark\\
ALOHA\cite{aloha} & \xmark & Puppeteering & Joint space & \cmark   & \cmark & None & \cmark\\       
Gello\cite{gello} & \xmark & Puppeteering & Joint space & \cmark   & \cmark & Passive spring mechanism & \cmark\\

Legato\cite{seo2024legato} & \cmark & Handheld & Task space & \xmark  & \cmark & None & \xmark\\
DexCap\cite{wang2024dexcapscalableportablemocap} & \cmark & Handheld & Task space & \xmark   & \xmark & Physical hand–object contact & \xmark\\
DexWild\cite{tao2025dexwild} & \cmark & Handheld & Task space & \xmark   & \cmark & Physical hand–object contact & \xmark\\
DexPilot\cite{DexPilot} & \xmark & Vision & -- & \xmark & \xmark & None & \cmark \\
Hato\cite{hato} & \xmark & VR & Joint space & \xmark & \cmark & None & \cmark \\
Song et al\cite{Song_2020} & \cmark & Handheld & Task space & \xmark  & \cmark & None & \xmark\\
DemoAT\cite{Young_2020} & \cmark & Handheld & Task space & \xmark  & \cmark & Passive trigger resistance & \xmark \\
ViViDex\cite{ViViDex} & \cmark & Vision & Task Space & \xmark  & \xmark & Physical hand–object contact & \xmark \\
ORION\cite{zhu2024visionbased} & \cmark & Vision & Graph-Based & \xmark  & \cmark & Physical hand–object contact & \xmark \\
\textbf{TRIP-Bag (Ours)} & \cmark & Puppeteering & Joint space & \cmark   & \cmark & Tracking-error feedback & \cmark \\
\bottomrule
\end{tabularx}
\end{table*}

The suitcase design allows deployment in everyday scenarios such as kitchens, workshops, and offices, enabling the study of manipulation tasks in naturalistic settings.
A puppeteer-based leader interface provides a direct mapping between operator and robot motion, reducing the embodiment gap, while the compact, plug-and-play hardware stack enables setup in a couple of minutes. The features of our system are a step toward scalable data collection for robot learning research.

Using our proposed system, we collected \sy{1238} demonstrations within 22 diverse environments. Fig. \ref{fig:main} illustrates the setup process of the data collection device along with examples of the environments in which data was gathered. We validated the collected dataset by training a manipulation policy, confirming the practicality of our proposed system as a portable data collection system.

The contribution of this paper summarizes as follows:
\begin{itemize}
    \item Teleoperation Anywhere: A fully portable teleoperation system compact enough to fit within a standard commercial suitcase, enabling researchers to easily transport and deploy the platform across diverse environments.
    \item \nm{Rapid} Deployment: A low-effort deployment pipeline, achieving high-fidelity teleoperation data collection within minutes of setup.
    \item Hardware Validation: Experiments showcasing the practicality of the proposed system in data collection and manipulation scenarios, highlighting its potential to accelerate the development of learning-based robot policies.
    
\end{itemize}

\begin{figure*}[t]
  \centering
\vspace*{10pt}  \includegraphics[width=1.0\textwidth]{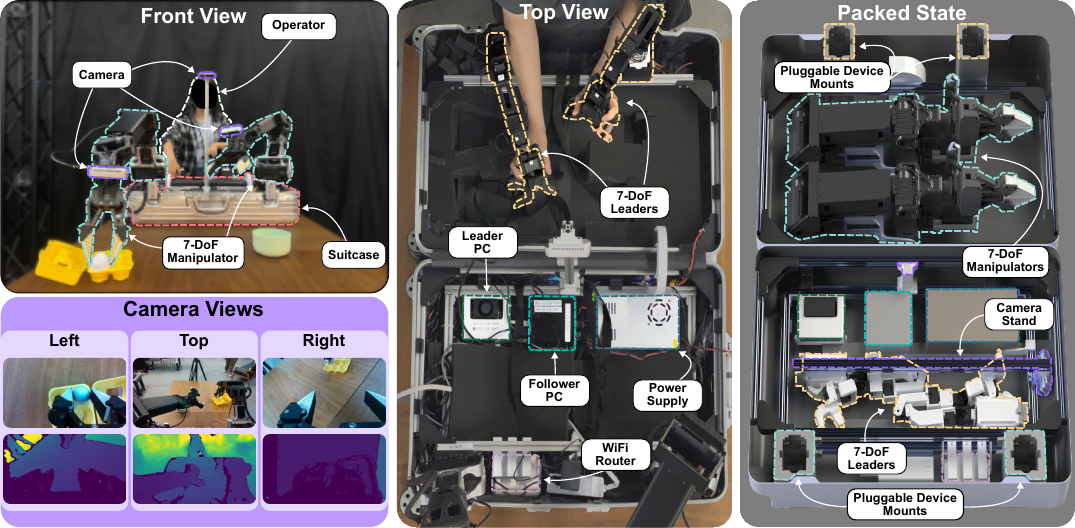}
  \caption{The breakdown of the hardware components and view of the cameras.}
  \label{fig:system-design}
\end{figure*}

\section{RELATED WORK}
\label{section:related_works}
Recent works have explored extracting demonstrations from in-the-wild videos using hand-pose estimator techniques\cite{ViViDex,Welschehold_2016,EgoMimic, zhu2024visionbased}, enabling large-scale, visually diverse data collection. However, noisy perception and inaccurate pose estimates hinder the recovery of precise, low-level motor commands necessary for reliable control.

Teleoperation with direct mapping has emerged as a more reliable alternative, where human-in-the-loop control produces high-quality demonstrations for downstream learning \cite{chi2024diffusionpolicy, aloha, chen2025tool}. A common approach uses virtual and augmented reality (VR/AR) interfaces to map human hand pose to robot end-effector motion, but this method remains sensitive to kinematic constraints and singularities due to morphological mismatches. Recently, to overcome the such mismatches, there have been works that utilize the puppeteering framework \cite{gello, aloha, fu2024mobile}. These systems employ a one-to-one joint mapping between the puppeteering device and the manipulator, enabling intuitive control and the collection of high-fidelity data without embodiment mismatch. While these teleoperation systems \nm{enable data collection of fine grained and long horizon manipulation tasks}, they are often constrained to a single environment making it difficult to gather in-the-wild data at scale. 

A common way to move beyond laboratory-bound setups is to use portable data collection devices \cite{Chi_2024,ha2024umilegs,seo2024legato,wang2024dexcapscalableportablemocap}. These systems typically employ either instrumented, hand-held grippers or wearable mocap gloves that track hand and finger kinematics with high spatial resolution. By decoupling data collection from fixed multi-camera rigs, they enable rapid, in-the-wild demonstration gathering at scale and with broad scene diversity. However, a persistent embodiment gap remains: human kinematics, compliance, and contact strategies often diverge from those of robot kinematics. As a result, substantial post-processing and retargeting, which includes calibration, time warping, and smoothing, are typically required to ensure robot-specific execution, safety, and dynamic feasibility. These pipelines are further challenged by sensing noise, drift/occlusion in wearable tracking, limited observability of object pose, and contact state.

\sy{As summarized in Table \ref{table:Compare}, TRIP-Bag bridges the gap between portable handheld interfaces and full joint-matched puppeteering systems. To the best of the authors knowledge it is the only approach that integrates portability, direct embodiment, calibration-free operation, and full proprioceptive state logging, enabling precise data collection through explicit joint mapping.}




\section{System Design}
To enable portable, easy-to-deploy, and efficient data collection, we designed TRIP-Bag with the following considerations.
\begin{enumerate}
    \item \textbf{Form Factor: } The system should maintain a compact and lightweight profile, making it easy to transport and deploy in diverse environments. 

    \item \textbf{Plug-and-Play Capabilities: } To reduce the burden of wiring, calibration, and setup, the system should allow for rapid assembly and disassembly. 

    \item \textbf{Intuitive Operation:} The teleoperation interface should be straightforward and require minimal training, allowing users to control the robot naturally and collect demonstrations with high fidelity. 
\end{enumerate}

\label{section:system}
\subsection{Hardware}
The teleoperation system utilizes a commercial suitcase as its housing. 
The check-in suitcase functions not only as a convenient means of global transport but also as an integrated deployment unit (Fig. \ref{fig:system-design}). The complete teleoperation system weighs 29.8 kg, complying with the standard airline over-weight allowance for check-in luggage.

We adopt a puppeteering teleoperation framework that provides an intuitive interface and facilitates seamless coordination between the operator and the follower robot. The leaders follow a scaled kinematic mapping to the manipulators, consistent with prior methods \cite{aloha, PAPRLE, gello}, enabling direct joint-level control and ensuring demonstrations with high kinematic fidelity. A single camera stand provides a stable top-down view from above the suitcase ensuring a consistent field of view during each session. 

\begin{figure}
    \centering
    \includegraphics[width=1\linewidth]{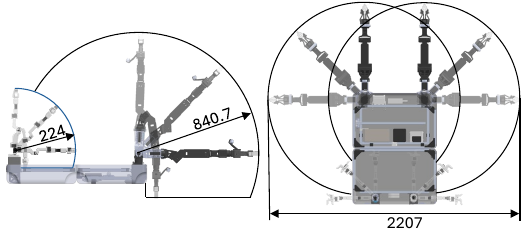}
    \caption{Workspace of TRIP-Bag. All dimensions are in mm.} 
    \label{fig:workspace}
\end{figure}

To achieve rapid deployment, both the leaders and the manipulators are mounted using a pluggable interface \cite{papras} (Fig. \ref{fig:system-design}). This design allows the manipulators to become operational within seconds, while also enabling them to fold into a compact form that fits neatly inside the suitcase when not deployed. As the manipulator workspace (Fig. \ref{fig:workspace}) is constrained by the leaders, the leader mounts are raised relative to the followers' to enable the followers to reach up to 100 mm below the suitcase bottom. \sy{The suitcase is reinforced with an aluminum frame to prevent deformation and maintain stability during operation. During transport, the components are surrounded by foam to protect against impact from handling.} 
Fig. \ref{fig:system-design} illustrates both the packed and deployed configurations of the system. A summary of all the components, and dimensions are provided in Table \ref{table:hw_components}.

The system relies on an external outlet for power distribution to two PCs (a leader PC and a follower PC), a portable Wi-Fi router, and a power supply, which in turn powers the pluggable manipulators and leaders.
\sy{The use of external power ensures full compliance with customs and airline safety regulations for the international deployment of the system. While the current configuration prioritizes regulatory compliance and long, uninterrupted operation, the design is fully capable of being converted to battery power for situations where portability is essential.}


\begin{table}[t]
\vspace*{5pt}
\caption{System Specifications}
\label{table:specifications}
\centering
\scriptsize
\begin{tabularx}{\columnwidth}{l l X}
\toprule
\textbf{Submodule} & \textbf{Specification} & \textbf{Details} \\
\midrule
\vspace{0.3em}
Suitcase & Weight & 14.4 kg \\\vspace{0.3em}
         & Dimension & 690 $\times$ 440 $\times$ 275 mm \\\vspace{0.3em}
         & Leader PC & Nvidia Jetson Orin Nano \\\vspace{0.3em}
         & Follower PC & Intel NUC10i3FNK \\\vspace{0.3em}
         & Wifi Router & Opal Wireless Travel Router \\\vspace{0.3em}
         & Power Supply & 24 V / 15 A \\
\midrule

PAPRAS (each) \vspace{0.3em}
         & Weight & 6.9 kg \\ \vspace{0.3em}
         & Length & 794 mm \\ \vspace{0.3em}
         & \#DoFs & 7 \\  \vspace{0.3em}
         & Payload & 2.5 kg \\ 

         & Actuators & DYNAMIXEL \\
         & & PH54-200-S500-R (200W) \\
         & & PH54-100-S500-R (100W) \\ \vspace{0.3em}
         & & PH42-020-S300-R (20W) \\ \vspace{0.3em}
         & Camera & Intel RealSense\texttrademark~D435 \\
         & Gripper & RH-P12-RN (Parallel Jaw) \\
\midrule

Pluggable Leader (each) \vspace{0.3em}
         & Weight & 0.46 kg \\\vspace{0.3em}
         & Length & 635 mm \\\vspace{0.3em}
         & \#DoFs & 7 \\
         & Actuators & DYNAMIXEL \\
         & & XL330-M288-T\\ 
         & &XL330-M077-T \\
\midrule

Top-Down Camera \vspace{0.3em}
         & Weight & 0.37 kg \\\vspace{0.3em}
         & Length & 685 mm \\\
         & Camera & Intel RealSense\texttrademark~D435 \\
\bottomrule
\end{tabularx}
\label{table:hw_components}
\end{table}


\subsection{Software}

\begin{figure}
    \centering
    \includegraphics[width=\linewidth]{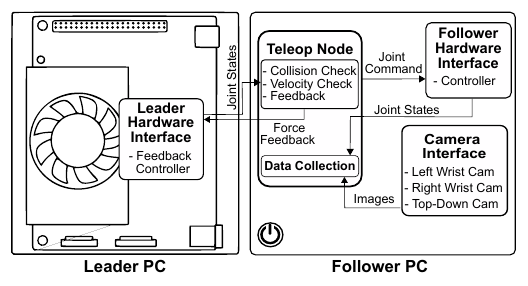}
    \caption{Overview of the software architecture, including device integration with the leader and follower PCs. }
    \label{fig:software}
\end{figure}

The overall teleoperation framework is based on PAPRLE \cite{PAPRLE} and Robot Operating System (ROS2).
Three nodes operate within teleoperation system, with the overall structure shown in Fig. \ref{fig:software}.
The follower hardware interface node handles motor actuation of the robot arms, subscribing to joint command topics from other nodes, and publishing the current joint states.
The leader hardware interface node manages the leader devices, also publishing their joint states and generating force feedback to help the operator maintain natural postures during teleoperation.
Finally, the teleoperation node subscribes to the leaders’ joint states and translates them into corresponding joint commands for the robot arms.
During this translation, the framework performs real-time self-collision checks and stops if a potential collision is detected, ensuring safe teleoperation for both the operator and the hardware. 
In addition, the teleoperation node also generates feedback signals  to guide the operator in situations where the follower cannot track the leader, such as obstacle collisions or joint limits. 
To collect data, the teleoperation node also subscribes to the follower joint states through ROS2 and subscribes to images through \texttt{pyrealsense2} API. 
During data collection sessions, the images and joint states are recorded as observation information, and corresponding joint commands are recorded as action information.

\section{Data Collection}
\label{section:data_collection}
With the proposed system, a user can collect manipulation datasets in diverse environments. In this section, we outline the setup procedure for data collection and detail the overall data collection process.
\begin{figure*}[t!]
    \centering
    \begin{subfigure}[t]{0.695\textwidth}
        \centering
        \includegraphics[width=1.0\textwidth]{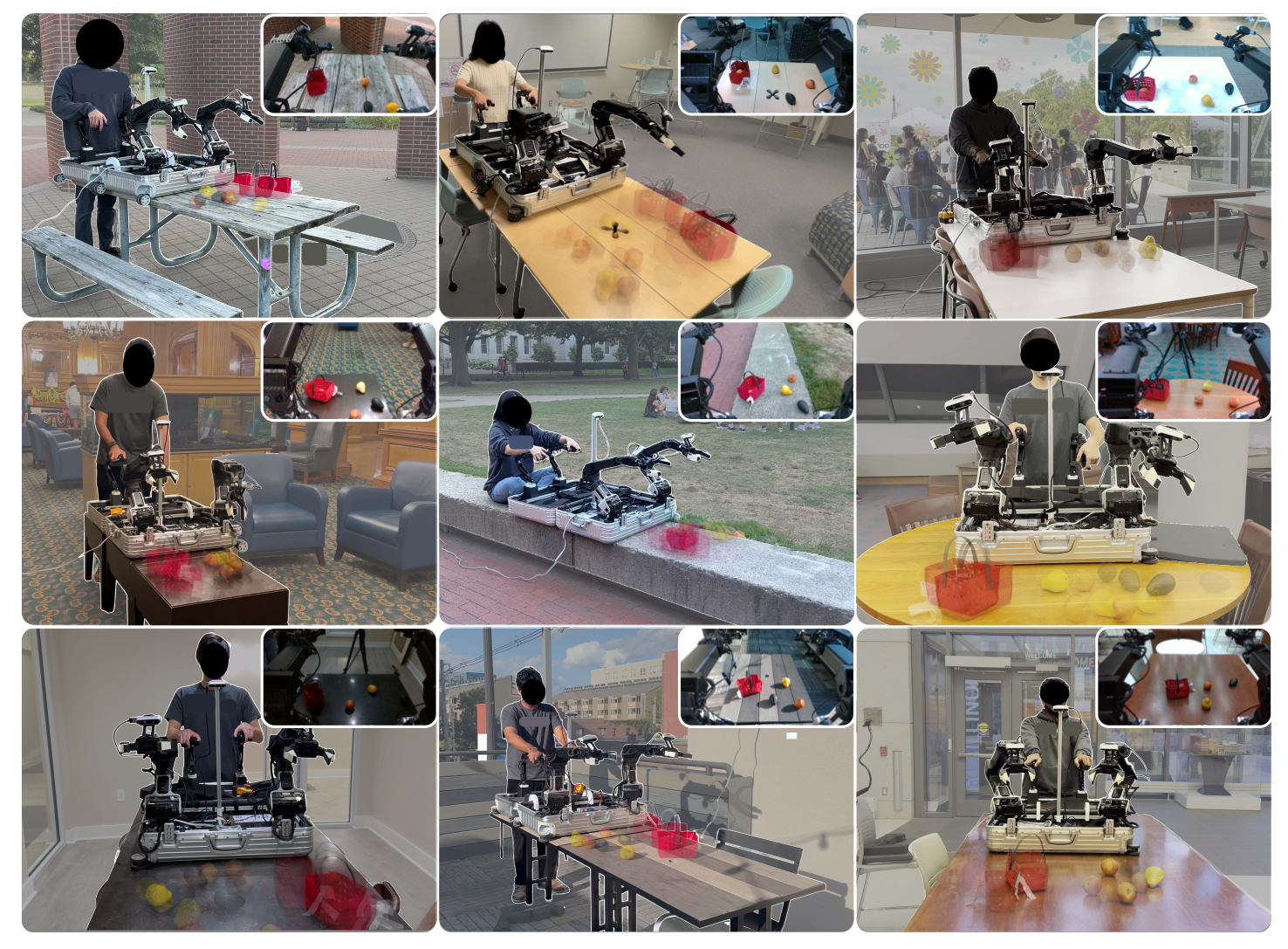}
        \caption{Example locations of collected data.}
        \label{fig:data-collection}
    \end{subfigure}%
    ~ 
    \begin{subfigure}[t]{0.265\textwidth}
        \centering
        \includegraphics[width=1.0\textwidth]{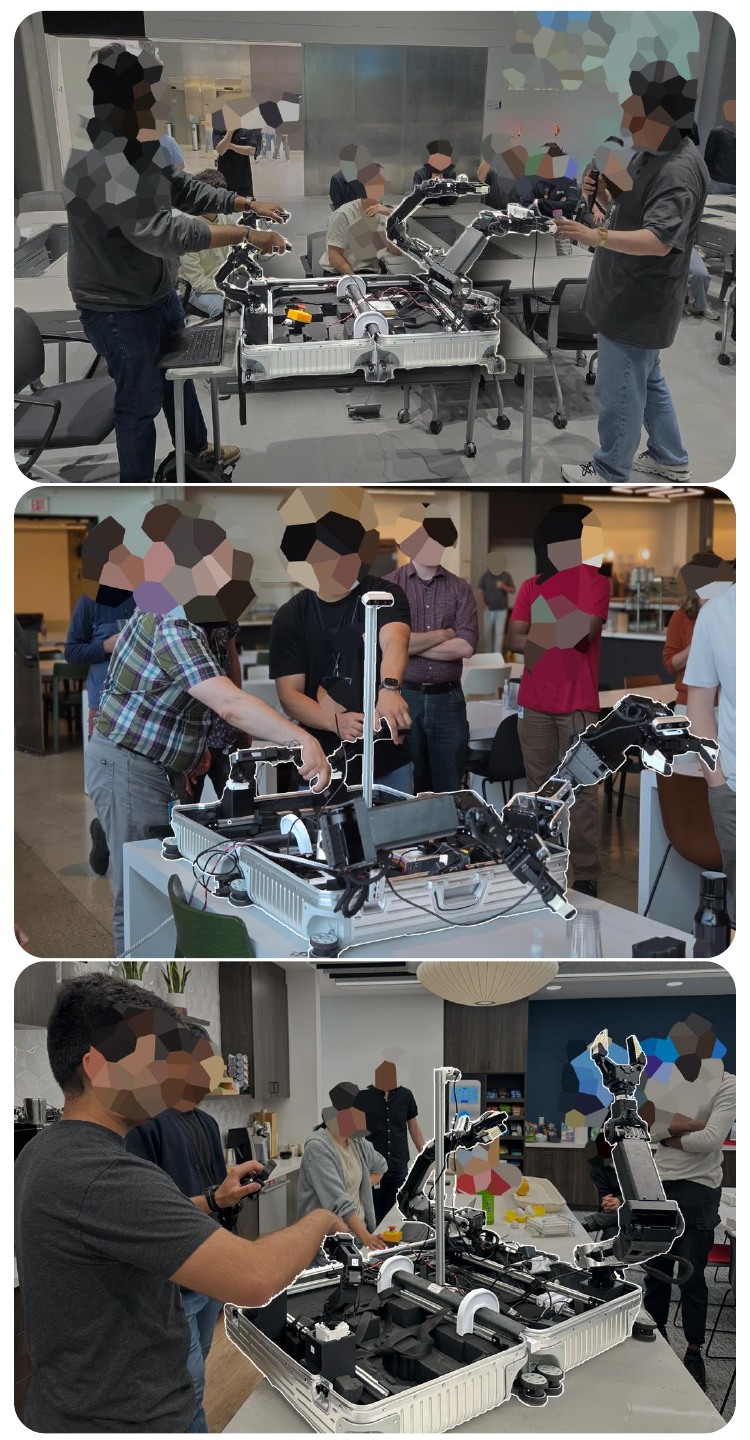}
        \caption{Demonstrations abroad.}
        \label{fig:demos}
    \end{subfigure}
    \caption{Range of deployment locations where the bag has been used for data collection and non-expert operation.}
\end{figure*}
\begin{figure*}[t]
  \centering
  \includegraphics[width=1.0\textwidth]{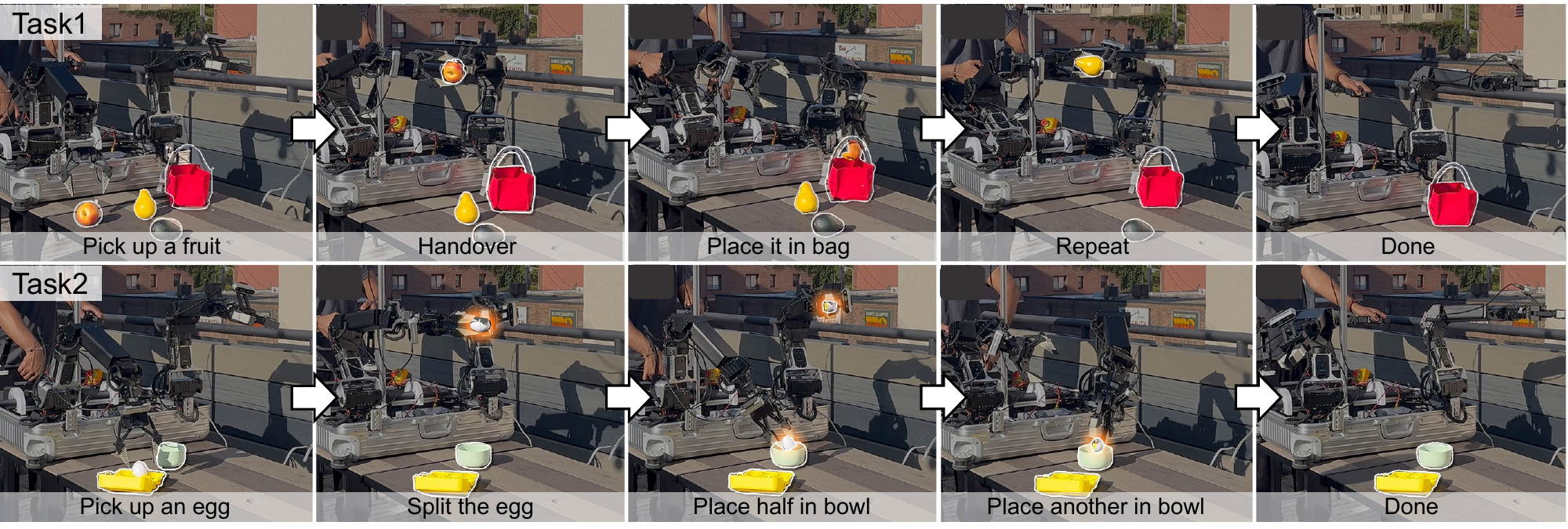}
  \caption{Step-by-Step Procedure for Target Tasks.}
  \label{fig:task}
\end{figure*}
\subsection{Setup}
The setup process begins by connecting the power and turning on the Follower PC.
Then, the camera stand is mounted, and the top-down camera is connected to the Follower PC.
The leader devices are then plugged in, followed by launching the leader control node. Next, the robots are connected, and the robot control node is launched. Afterward, the wrist cameras are connected, and the teleoperation node is launched on the Follower PC. Once these steps are completed, the system is ready for data collection. Fig. \ref{fig:main} shows the overall process of setup. 

The supplementary video demonstrates the entire setup process in detail. On average, the setup takes an expert operator less than five minutes (average 200 seconds), from opening the suitcase to running the first teleoperation session.
This simple and fast setup enables flexible collection of robot manipulation data across diverse environments as shown in Fig. \ref{fig:data-collection}.

\subsection{Data Collection Process}
The data collection process begins with the initialization of the teleoperation node, after which the manipulators move to a predefined ready pose, waiting for a start signal from leader devices. A session is initiated when the operator grasps both leader grippers for a specified duration (one second). Upon receiving the start signal, the followers transitions to the initial configuration given by the leaders initial state and then begins following their commands, allowing the operator to perform the target task. 

Each session corresponds to a single demonstration, defined as the full-horizon trajectory required to complete the target task. During the session, cameras are streamed at 30hz, while joint states (position, velocity, and effort) are updated at 125hz. All data are synchronized and recorded at 50 hz.  

To terminate the session, the operator grasps both leader grippers within a user designated ending zone for a specified duration (one second), sending the stop signal. The teleoperation node then stops recording and returns the robot to the ready pose. This procedure, adopted from PAPRLE \cite{PAPRLE}, allows the operator to collect multiple demonstrations with minimal setup between demonstrations.



\section{Experiments}
\label{section:experiments}

The portability of our system allows users to collect datasets in diverse environments. 
We have collected demonstrations in multiple different environments (22 different environments, totaling \sy{1238} demonstrations), mainly for the two tasks described in Sec. \ref{subsection:target_tasks}. Fig. \ref{fig:data-collection} shows the environments where datasets were collected and the corresponding top-down camera images for each environment.

We also collected 200 demonstrations from 10 different non-expert users, to analyze the usability and diversity of data \nm{collected} using this system.  
Using the \nm{full} dataset, we trained a manipulation policy to verify the quality of the data and to validate the practicality of the proposed system.

\begin{figure*}[t!]
    \vspace*{5pt}

    \centering
    \begin{subfigure}[t]{0.35\textwidth}
        \centering
        \includegraphics[width=0.97\textwidth]{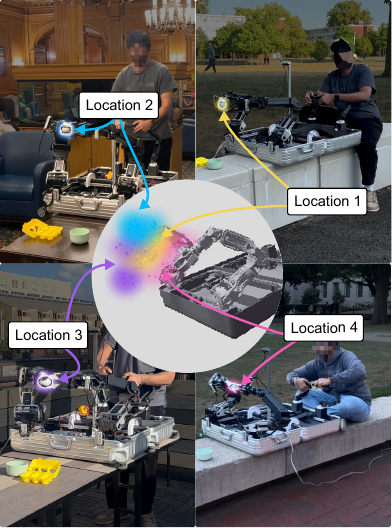}
        \caption{Data collected by one user in various environments. }
        \label{fig:egg_comparison_sankalp}
    \end{subfigure}%
    ~ 
    \begin{subfigure}[t]{0.65\textwidth}
        \centering
        \includegraphics[width=0.97\textwidth]{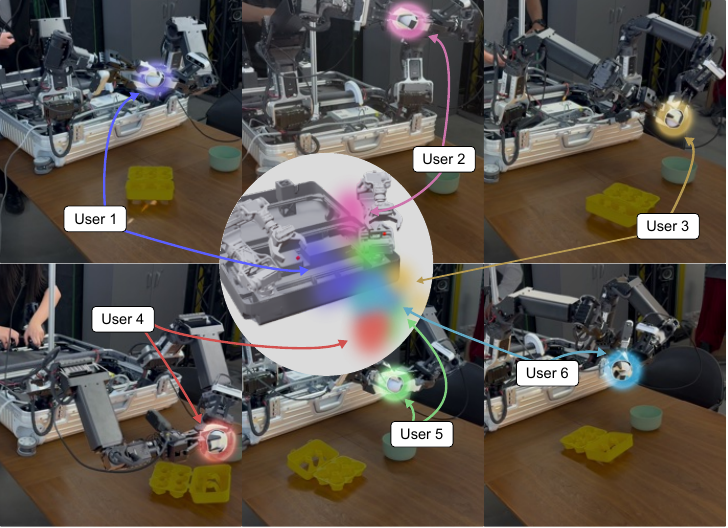}
        \caption{Data collected by various user in one environment. }
        \label{fig:egg_comparison_kids}
    \end{subfigure}
    \caption{Set of interaction points from Task 2. Each colored cloud in the middle represents each set of the interaction points by categorization (For (a), based on location. For (b), based on users). Pictures shows the example interaction point from each set.}
\end{figure*}
\subsection{Target Tasks}
\label{subsection:target_tasks}
\sy{We selected two bimanual tasks \nm{representing two extremes of bimanual teleoperation} to demonstrate our system. The first task requires high spatial and sequential bimanual coordination through repeated pick, handover, and place motions. The second task focuses on fine-grained precision and simultaneous collaborative motions for the system to achieve success. Fig. \ref{fig:task} illustrates the overall process for each task. The supplementary video demonstrates example executions of these tasks, as well as additional teleoperated tasks that highlight the capabilities of TRIP-Bag. Expanding the dataset to include more tasks is reserved for future work.}

\subsubsection{Task 1: Fruit Collecting}
In this task, three toy fruits (selected from 2 apples, 1 pear, and 1 avocado) and a basket are placed in the workspace. The fruits are placed closer to the right arm, and the basket is placed near the left arm. Since it is difficult for the left arm to reach the fruits directly, the right arm must pick them up one by one and hand them over to the left arm. The left arm then places each fruit into the basket.

This task requires careful handover, making coordination between the two arms essential. Furthermore, because each fruit has a different shape (apple, pear, avocado), the operator must adapt their grasp strategy for each fruit to make the handover easier.
The task is considered successful when all three fruits are placed in the basket. If any fruit falls during the process, the attempt is regarded as a failure.

\subsubsection{Task 2: Egg Cracking}
In this task, a toy egg is placed on a carton near the right arm. The egg has a seam along its middle, allowing it to split into two halves. The right arm grasps the egg and brings it into position for both arms to crack it open together. Each half of the egg is then placed into a bowl.
This task requires collaborative motion of both arms to perform a tightly coordinated action. 
The egg is relatively small and slippery, so the user must carefully regulate grasping force: gripping too loosely may drop the egg, while gripping too tightly may cause it to slip out of the gripper.
The task is considered successful when both halves are placed into the bowl. During the demonstration, regrasping is allowed during the cracking step; however, if any part of the egg is dropped, the trial is considered a failure. 

%


\subsection{Portability } \label{subsection:portability_usability}

\begin{figure}[t]
    \centering \includegraphics[width=\columnwidth]{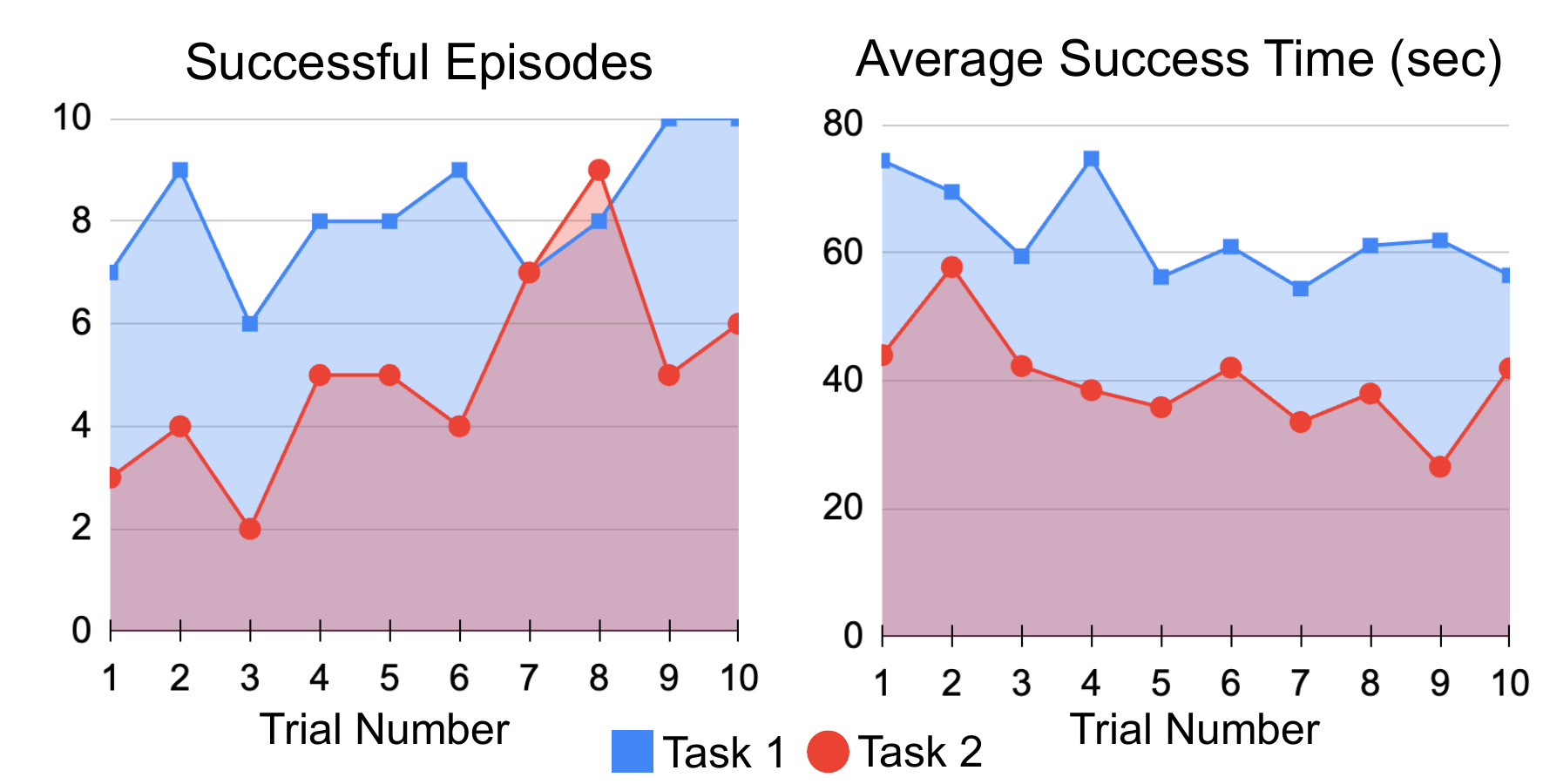} 
    \caption{The number of successful episodes and average success time from the non-expert users.}
    \label{fig:kids-stats}
\end{figure}

We evaluated the setup time of the proposed system across eight different environments and with multiple users.
On average, the system could be fully prepared and ready for the first teleoperation session in 200 seconds. 

The accompanying video provides the setup recordings from all eight environments.
This demonstrates that the system can be easily carried and deployed in real-world settings, enabling data collection within minutes of arrival and supporting rapid dataset collection in the wild.
The system was also transported overseas as standard checked luggage and used for demonstrations, as illustrated in Fig.~\ref{fig:demos}, highlighting its portability.
\begin{figure*}
    \centering \includegraphics[width=\textwidth]{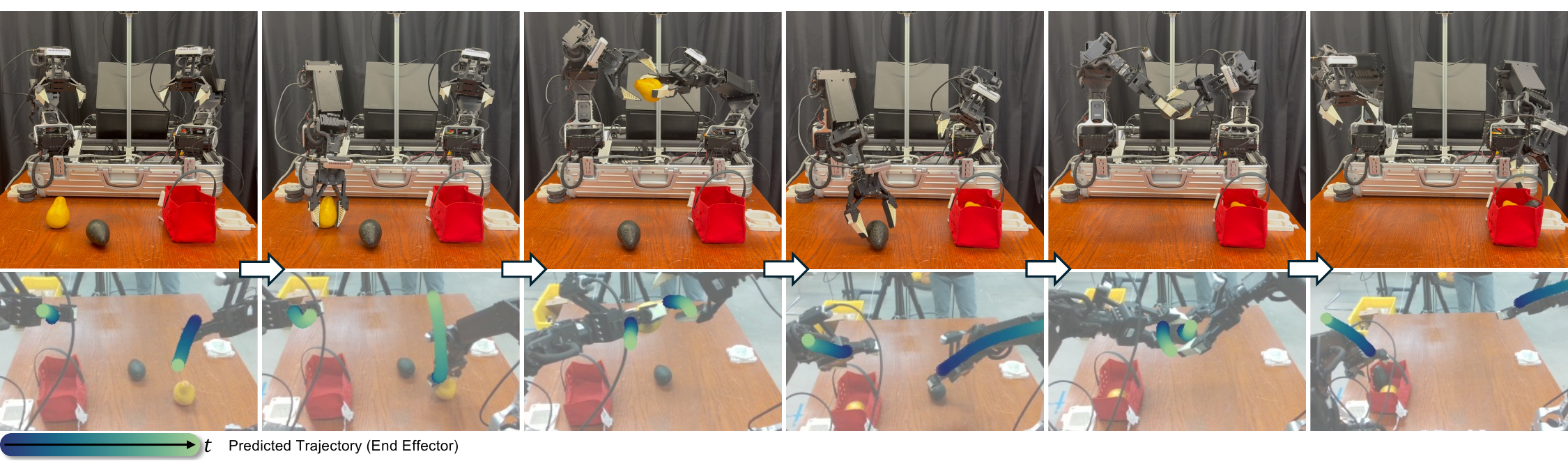} 
    \caption{Example Policy Rollouts for Task 1. Bottom row shows the predicted trajectory of end-effector from trained policy, calculated using forward kinematics models}
    \label{fig:learning_fruits}
\end{figure*}

\begin{figure*}
    \centering \includegraphics[width=\linewidth]{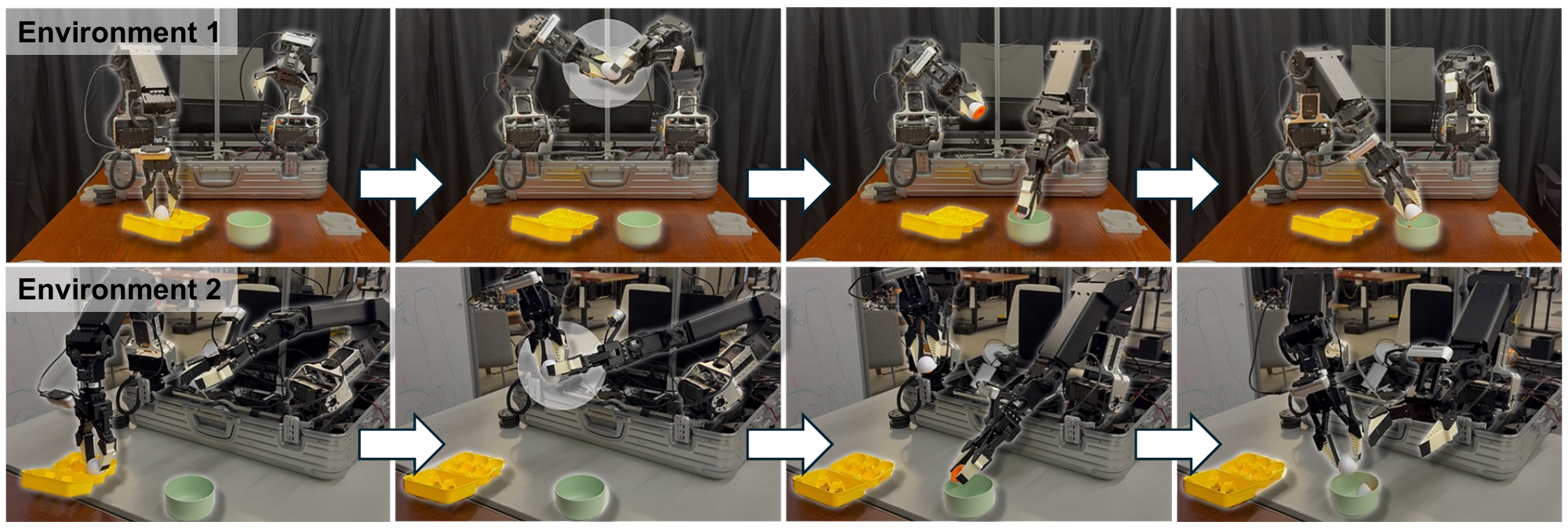} 
    \caption{Example Policy Rollouts for Task 2.}
    \label{fig:learning_eggs}
\end{figure*}
\subsection{Usability} 
\label{subsection:usability}
To evaluate the ease of use of our system, we also collected a dataset from non-expert users with no prior experience of teleoperation systems.
We report the analyses from the dataset which is collected by a total of 10 non-expert users. 

Each user first watched a three minute instructional video explaining how to operate the system and the goals of the two tasks described in Sec. \ref{subsection:target_tasks}, and was then given 10 trials per task.
Considering Task 2 (Egg Cracking) is relatively more difficult than Task 1 (Fruit Collecting), the users performed Task 1 first, followed by Task 2.
Fig. \ref{fig:kids-stats} summarizes the results, showing the number of users who succeeded across trials and the average completion time for successful demonstrations. 
Despite having no prior exposure to puppeteering or teleoperation systems, many users successfully completed Task 1 on their very first attempt. 
Over the course of the trials, all 10 participants succeeded in Task 1, and the average completion time decreased steadily, indicating fast familiarization with the system.
For Task 2, participants initially struggled due to the need for more precise force control and bimanual coordination. 
However, success rates improved and average completion times decreased over repeated trials.

These findings suggest that the proposed system is intuitive and easy to learn, even for users without prior teleoperation experience. 

\subsection{Data Diversity} \label{subsection:data_diversity}
As shown in Figure \ref{fig:data-collection}, we have collected demonstration data for two tasks in many different environments. 
These environmental variations introduce diversity in the image domain, resulting in similar workspace configurations but different backgrounds and overall visual appearances.
Interestingly, we found that these environmental differences affect not only the visual data but also the robot trajectories. 
For example, Fig. \ref{fig:egg_comparison_sankalp} illustrates the interaction points between the two arms during the egg-splitting phase of Task 2, from the same user. 
Despite being performed by the same user on the same task, the average interaction points shift depending on the environment in which the data was collected.
This is largely due to changes in the relative height and placement of the system in different locations, which alter the operator’s posture and lead to naturally different motion patterns.

We also observed diversity across users, even when operating in the same environment with identical task settings. 
By analyzing the Task 2 trajectories from \ref{subsection:usability}, we plotted the interaction points in Fig. \ref{fig:egg_comparison_kids}. 
Each participant exhibited a distinct interaction zone, reflecting individual differences in teleoperation style.
\nm{Such diversity is a valuable property of our system.} As prior work has shown that diverse datasets enable more robust manipulation policies \cite{data_diversity,khazatsky2024droid}, our portable teleoperation system can serve as a practical tool for collecting diverse demonstrations.

\subsection{Feasibility of Learning from Collected Data}

To validate that the collected dataset is suitable for learning, we trained baseline manipulation policies. 
For the policy architecture, we adopted the Action Chunking Transformer (ACT) \cite{aloha}, maintaining large parts of its standard design. 
Specifically, the input consists of three RGB-D images from the cameras together with the current joint states, while the output is a predicted trajectory of future joint states.
A separate ACT policy was trained for each task.
Training was conducted on two NVIDIA A40 GPUs per task, and inference was performed on the a laptop with NVIDIA RTX 4070 Laptop GPU, connected to the system.

The trained policies were able to complete the tasks, and example policy rollouts are shown in Fig. \ref{fig:learning_fruits} and Fig. \ref{fig:learning_eggs}. 
In Fig. \ref{fig:learning_fruits}, the robot is conducting Task 1: right arm successfully hands over fruits to the left arm, with the predicted trajectories shown at the bottom. 
Although the policies often fail to achieve precise grasps, we found that the robot repeatedly attempts to re-grasp objects after failures, indicating learned environment perception and persistence.
Similarly, Fig. \ref{fig:learning_eggs} illustrates rollouts for Task 2, where the robot executes different grasping poses during the egg-splitting phase. 
Supplementary video includes the examples of these policy executions.
While our goal is not to achieve state-of-the-art task performance, these rollouts demonstrate that the dataset collected with our system can support policy training and produce behaviors aligned with the tasks. We expect the proposed system to serve as a practical and portable platform for collecting diverse manipulation datasets and facilitating learning experiments in real-world environments.

\section{Conclusion}
\label{section:conclusion}
In this paper, we introduced TRIP-Bag, a portable teleoperation system for data collection. 
The system can be set up within minutes, allowing users to easily collect diverse robot manipulation demonstrations across a wide range of environments. 
We demonstrated its portability by collecting datasets in multiple settings and showed, through experiments with non-expert users, that the system is intuitive and easy to learn, thus lowering the entry barrier for broader adoption.
We further validated the effectiveness of the collected data by training manipulation policies, with results confirming that datasets obtained through our system provide a valuable resource for the robot learning community. \green{Going forward, we plan to collect more data across a wider set of tasks, which will help the system handle more diverse scenarios. With this expanded dataset, we can continue refining the learning policies, making them more robust and better aligned with the types of tasks we want to target.}








\bibliographystyle{IEEEtran}
\bibliography{reference.bib}

\end{document}